\documentclass[journal,preprints,submit,pdftex,moreauthors]{Definitions/mdpi}
\preto{\abstractkeywords}{\nolinenumbers}
\firstpage{1} 
\pubvolume{1}
\issuenum{1}
\articlenumber{0}
\pubyear{2022}
\copyrightyear{2022}
\datereceived{} 
\dateaccepted{} 
\datepublished{} 
\hreflink{https://doi.org/} 

\usepackage{graphicx}
\usepackage{multirow}
\usepackage{adjustbox}
\usepackage{float}
\usepackage{color,soul}
\usepackage{todonotes}
\usepackage{listings}
\restylefloat{table}
\usepackage{xcolor}

\Title{Enriching Artificial Intelligence Explanations with Knowledge Fragments}

\TitleCitation{Enriching Artificial Intelligence Explanations with Knowledge Fragments}

\Author{Jo\v{z}e M. Ro\v{z}anec $^{1,2,3,\dagger}$* \orcidA{}, Elena Trajkova $^{4,}$ \orcidG{}, Inna Novalija $^{2,}$ \orcidB{}, Patrik Zajec $^{1,2,}$ \orcidC{}, Klemen Kenda $^{1,2,3,}$ \orcidD{}, Bla\v{z} Fortuna $^{3,}$ \orcidE{} and Dunja Mladeni\'{c} $^{2,}$ \orcidF{}}

\AuthorNames{Jo\v{z}e M. Ro\v{z}anec, Elena Trajkova, Inna Novalija, Patrik Zajec, Klemen Kenda, Bla\v{z} Fortuna and Dunja Mladeni\'{c}}

\AuthorCitation{Ro\v{z}anec, J.M.; Trajkova E.; Novalija I.; Zajec P.; Kenda K.; Fortuna B.; Mladeni\'{c} D.}

\address{%
$^{1}$ \quad Jo\v{z}ef Stefan International Postgraduate School, Jamova 39, 1000 Ljubljana, Slovenia\\
$^{2}$ \quad Jo\v{z}ef Stefan Institute, Jamova 39, 1000 Ljubljana, Slovenia\\
$^{3}$ \quad Qlector d.o.o., Rov\v{s}nikova 7, 1000 Ljubljana, Slovenia\\
$^{4}$ \quad University of Ljubljana, Faculty of Electrical Engineering, Tr\v{z}a\v{s}ka c. 25, 1000, Ljubljana, Slovenia}

\corres{Correspondence: joze.rozanec@ijs.si; (J.M.R.)}

\abstract{Artificial Intelligence models are increasingly used in manufacturing to inform decision-making. Responsible decision-making requires accurate forecasts and an understanding of the models' behavior. Furthermore, the insights into models' rationale can be enriched with domain knowledge. This research builds explanations considering feature rankings for a particular forecast, enriching them with media news entries, datasets' metadata, and entries from the Google Knowledge Graph. We compare two approaches (embeddings-based and semantic-based) on a real-world use case regarding demand forecasting.}


\begin{document}


\section{Introduction}\label{INTRODUCTION}

The industry is the part of the economy that is concerned with a highly mechanized and automated production of material goods \cite{lasi2014industry}. Since the beginning of industrialization, technological breakthroughs allowed to achieve increasing manufacturing efficiencies and led to paradigm shifts. The latest such shifts are known as Industry 4.0 and Industry 5.0 \cite{erro2019industry,maddikunta2021industry}, and aim to foster and increase the digitalization, networking, and automation of the manufacturing processes. Furthermore, these, in turn, enable tighter integration between physical and cyber domains (e.g., through cyber-physical systems \cite{lu2017cyber}, and digital twins \cite{shafto2010draft}), increased deployment of intelligence (e.g., through Artificial Intelligence solutions \cite{arinez2020artificial}), to achieve autonomy and automation. These changes are expected to shorten the development and manufacturing periods, increase the manufacturing efficiency and sustainability \cite{ghobakhloo2020industry}, and achieve greater flexibility. Furthermore, emphasis on making such technologies and applications safe, trustworthy, and human-centric is a crucial characteristic of the nascent Industry 5.0 paradigm \cite{martynov2019information,rovzanec2022human}.

The increasing digitalization and the democratization of Artificial Intelligence have enabled such models to automate and provide assistance on a wide range of tasks. Machine Learning models can learn from historic data to predict future outcomes (e.g., estimate future demand \cite{rovzanec2021automotive}) or perform certain tasks (e.g., identify manufacturing defects \cite{trajkova2021active} or automate manual tasks with robotic assistance \cite{bhatt2020expanding}). Nevertheless, it is crucial to establish a synergic and collaborative environment between humans and machines, where humans can flourish in developing their creativity while delegating monotonous and repetitive tasks to machines \cite{rovzanec2022human}. Furthermore, such collaboration requires transparency to develop trust in Artificial Intelligence applications. Research exploring the models' rationale behind the predictions and how the insights must be provided to the user as an explanation is known as Explainable Artificial Intelligence.

Insights obtained regarding the models' rationale behind a prediction can be enriched with domain knowledge to provide a better-contextualized explanation to the user \cite{dhanorkar2021needs,dragoniknowledge}. To encode domain knowledge, researchers have resorted to using semantic technologies, such as ontologies and knowledge graphs. In order to integrate semantic technologies into the explanation crafting process, the structure of the explanation must be defined, and a procedure established to (a) extract semantic meaning from the explanation, (b) use such information to query external sources (e.g., open knowledge graphs), and (c) enrich the explanation with new information and insights obtained from the external sources.

In this research, we extend the work performed at \cite{rovzanec2022knowledge} and \cite{rovzanec2021xai}. Given a set of demand forecasts, a mapping between features and concepts that define an ontology and a hierarchy of concepts, and a set of keywords associated with each concept, we recourse to a wikification process to extract wiki concepts based on the keywords mentioned above. The wiki concepts are then used to query external sources (e.g., open knowledge graphs or open datasets) and rank query results based on their semantic similarity computing the Jaccard distance. Our approach was tested in the domain of demand forecasting and validated on a real-world case study, using models we developed as part of the European Horizon 2020 projects FACTLOG\footnote{https://www.factlog.eu/} and STAR\footnote{http://www.star-ai.eu/}. We evaluate our results through two metrics (precision@K and Ratio of Diverse Entries (RDE@K)) to assess how precise and diverse the knowledge fragments are.

The rest of this paper is structured as follows: Section~\ref{RELATED-WORK} presents related work, Section~\ref{USE-CASE} describes the use case we used and the implementation we followed to test our concept, Section~\ref{RESULTS-AND-EVALUATION} provides the results we obtained and their evaluation. Finally, in Section~\ref{CONCLUSION}, we provide our conclusions and outline future work.

\section{Related Work}\label{RELATED-WORK}
\subsection{Industry 4.0 and Industry 5.0}
Industry 4.0 was coined in 2011 by the German government initiative aimed at developing advanced production systems to increase the productivity and efficiency of the manufacturing industry. The initiative was soon followed by national initiatives across multiple countries, such as the USA, UK, France, Canada, Japan, and China, among others \cite{majstorovic2019industry,bogoviz2019comparative,raj2020barriers}. Industry 4.0 was conceived as a technological revolution adding value to the whole manufacturing and product lifecycle. Part of such a revolution is the concept of a smart and integrated supply chain, which aims to reduce delivery times and reduce information distortion across suppliers and manufacturers \cite{frank2019industry}. The benefits mentioned above are achieved by enhancing the demand forecasting and optimizing the organization and management of materials, suppliers, and inventory \cite{ghobakhloo2018future,zheng2021applications}. To that end, digital twins of existing processes can be created to simulate, test \textit{what-if} scenarios, and enhance them without disruption of the physical operations \cite{qi2018digital,rovzanec2022actionable}.

On top of the aforementioned integrated supply chain, the Industry 4.0 paradigm emphasizes the redesign of the human role in manufacturing, leveraging new technological advancements and capabilities \cite{frank2019industry}. This aspect is further evolved in Industry 5.0. Industry 5.0 is a value-driven manufacturing paradigm that underscores the relevance of research and innovation to develop human-centric approaches when supporting industry operation \cite{xu2021industry}. Human-centricity in manufacturing must take into account the skills that are unique to the human workers, such as critical thinking, creativity, and domain knowledge while leveraging machine strengths (e.g., high efficiency in performing repetitive tasks \cite{nahavandi2019industry,demir2019industry,maddikunta2021industry}). Human-centricity can be realized through (a) a systemic approach focused on forging synergic, two-way relationships between humans and machines, (b) the use of digital twins at a systemic level, and (c) the adoption of artificial intelligence at all levels \cite{Industry50,rovzanec2022human}.

Artificial Intelligence can be essential to achieving human-machine collaboration. In particular, Active Learning and Explainable Artificial Intelligence can be used to complement each other. Active Learning is the sub-field of Artificial Intelligence, concerned with retrieving specific data and leveraging human knowledge to satisfy a particular learning goal. On the other hand, Explainable Artificial Intelligence is the sub-field of Artificial Intelligence concerned with providing insights into the inner workings of a model regarding its outcome so that the user can learn about its underlying behavior. This way, the human can act as a teacher to Artificial Intelligence models and learn from them through Explainable Artificial Intelligence. This two-way relationship can lead to a trusted collaboration \cite{weitz2019you,honeycutt2020soliciting}.

\subsection{Demand Forecasting}
Demand forecasting is a key component of manufacturing companies. Precise demand forecasts allow to set correct inventory levels, price the products and plan future operations. Any improvements in such forecasts directly translate to the supply chain performance \cite{moroff2021machine}. Demand depends on characteristics that are intrinsic to the product \cite{purohit2001effect} (e.g., elasticity or configuration), and external factors \cite{callon2002economy} (e.g., particular sales conditions). 

Researchers developed multiple schemas to classify demand according to its characteristics. Among them we find the ABC inventory classification system \cite{teunter2010abc}, the XYZ analysis \cite{scholz2012integration}, and the quadrant proposed by Syntetos et al. \cite{syntetos2005categorization}. The ABC inventory classification system classifies the items based on their decreasing order of annual dollar volume. The XYZ analysis classifies items according to their consumption patterns: (X) constant consumption, (Y) fluctuating (usually seasonal) consumption, and (Z) irregular consumption. Finally, Syntetos et al. divide demand into four quadrants based on demand size and demand occurrence variability. It has been recognized that Artificial Intelligence models can provide accurate demand forecasts based on past demand and complementary data. Different methods are appropriate for demands with different characteristics \cite{rovzanec2021reframing}. Demand forecasting models are usually framed as time series forecasting problems using supervised regression models or specialized models to learn patterns and forecast future values. Research related to the automotive industry has reported using complementary sources such as unemployment rate \cite{bruhl2009sales}, inflation rate \cite{vahabi2016sales} or Gross Domestic Product \cite{ubaidillah2020study}, and a variety of algorithms, such as the Multiple Linear Regression \cite{dwivedi2013business}, Support Vector Machine \cite{wang2020making}, and Neural Networks \cite{chandriah2021rnn}.

The ability to leverage a wide range of complementary data sources is a specific advantage of such models against humans in light of their visual working memory, short-term memory, and capacity to process variables \cite{halford2005many}. Furthermore, Artificial Intelligence models avoid multiple cognitive biases to which humans are prone to \cite{barnes1984cognitive}. Nevertheless, planners must approach such forecasts with critical thinking since they hold responsibility for decisions based on such forecasts. They must understand the models' rationale behind the forecast \cite{adadi2018peeking,arrieta2020explainable,confalonieri2021historical}, take into account information that could signal adjustments to the forecast are needed and make such adjustments when needed \cite{davydenko2010judgmental,davydenko2013measuring,alvarado2017expertise}. Furthermore, while transparency regarding the models' underlying rationale can be in some cases required by law \cite{goodman2017european} (e.g., the General Data Protection Regulation(GDPR) \cite{GDPR} or the Artificial Intelligence Act \cite{AIACT}), it also provides a learning opportunity, which is key to employees' engagement \cite{anitha2014determinants}.

\subsection{Explainable Artificial Intelligence}
Explainable Artificial Intelligence in a sub-field of Artificial Intelligence research, concerned with how the models' behavioral aspects can be translated into a human interpretable form to understand causality, enhance trustworthiness, and develop confidence \cite{emmert2020explainable,schwalbe2021xai}. Techniques are usually classified based on their complexity (degree of interpretability), their scope (global or local), and whether they are model-agnostic \cite{adadi2018peeking}. Regarding their degree of interpretability, models are usually considered black-box (opaque) or white-box (transparent). The source of models' opacity can be due to the inherent properties of the algorithms, their complexity, or due to the explicit requirement to avoid exposing its inner workings (e.g., trade secret), \cite{chan2021explainable,muller2021deep}.

Demand forecasting can be framed as a regression problem, and thus explainability techniques developed for this kind of supervised learning can be used to unveil the models' inner workings. One such technique is LIME \cite{ribeiro2016should} (Local Interpretable Model-agnostic Explanations), which provides a model-agnostic approach to estimate features' relevance for each forecast. LIME creates a linear model to approximate the model's behavior at a particular forecasting point. Then, it estimates the features' relevance by measuring how much predictions change upon features' perturbation. Similar approaches were later developed to ensure explanations were deterministic \cite{zafar2019dlime}, or take into account non-linear relationships between the features \cite{hall2017machine,sokol2020limetree}. Other frequently cited methods are SHAP \cite{lundberg2017unified,strumbelj2010efficient} (SHapley Additive exPlanations, which leverages coalitional game theory to estimate the contributions of individual feature values), LACE \cite{pastor2019explaining} (Local Agnostic attribute Contribution Explanation, which leverages SHAP and LIME to provide an explanation through local rules), LoRE \cite{guidotti2018local} (LOcal Rule-based Explanation, which crafts an explanation extracting a decision rule and a set of counterfactual rules), Anchors \cite{ribeiro2018anchors} (which establishes a set of precise rules that explain the forecast at a local level), and Local Foil Trees \cite{van2018contrastive} (which identifies a disjoint set of rules that result in counterfactual outcomes).

While understanding why the model issued a particular forecast is of utmost importance, explanations crafting can be guided using domain knowledge, enriched with complementary data or insights. An approach to enhance explanations given specific domain knowledge was developed by Confalonieri et al. \cite{confalonieria2021towards}, who demonstrated that decision trees built for explainability were more understandable if built considering domain knowledge. Enrichment with complementary data was researched by Panigutti et al. \cite{panigutti2020doctor}, where high fidelity to the forecasting model was achieved by enriching the explainability rules with semantically encoded domain knowledge. Semantic enrichment or recourse to graph representations to bind multiple insights was explored by several authors \cite{lecue2019thales,rovzanec2022knowledge,dragoniknowledge}. Finally, Rabold et al. \cite{rabold2019enriching} devised means to create enhanced, multimodal explanations by relating visual explanations to logic rules obtained through an inductive logic programming system.

While many metrics have been devised to assess the quality of the outcomes produced by explainability techniques \cite{lakkaraju2017interpretable,nguyen2020quantitative,rosenfeld2021better,amparore2021trust}, explanations must also be evaluated and measured from a human-centric perspective. Attention must be devoted to ensuring the explanations convey meaningful information to different user profiles according to their purpose \cite{samek2019towards,pedreschi2018open,el2019towards}, that they promote curiosity to increase learning and engagement \cite{hsiao2021roadmap}, and provide means to develop trust through exploration \cite{hoffman2018metrics}. Furthermore, it is desired that the explanations are actionable \cite{keane2020good,keane2021if}, and inform conditions that could change the forecast outcome \cite{verma2020counterfactual}. The aspects mentioned above are frequently evaluated through qualitative interviews and questionnaires, think-aloud, or self-explanations \cite{hsiao2021roadmap,mohseni2021multidisciplinary}. Methods such as tracking participants' eye movement patterns, measuring users' response time to an explanation, the quantification of answers provided, or the number of explanations required by the user to learn have also been proposed \cite{lage2018human,sovrano2021objective,hsiao2021roadmap,mohseni2021multidisciplinary}.

\section{Use case}\label{USE-CASE}

\begin{figure*}[ht!]
\centering
\includegraphics[width=\textwidth]{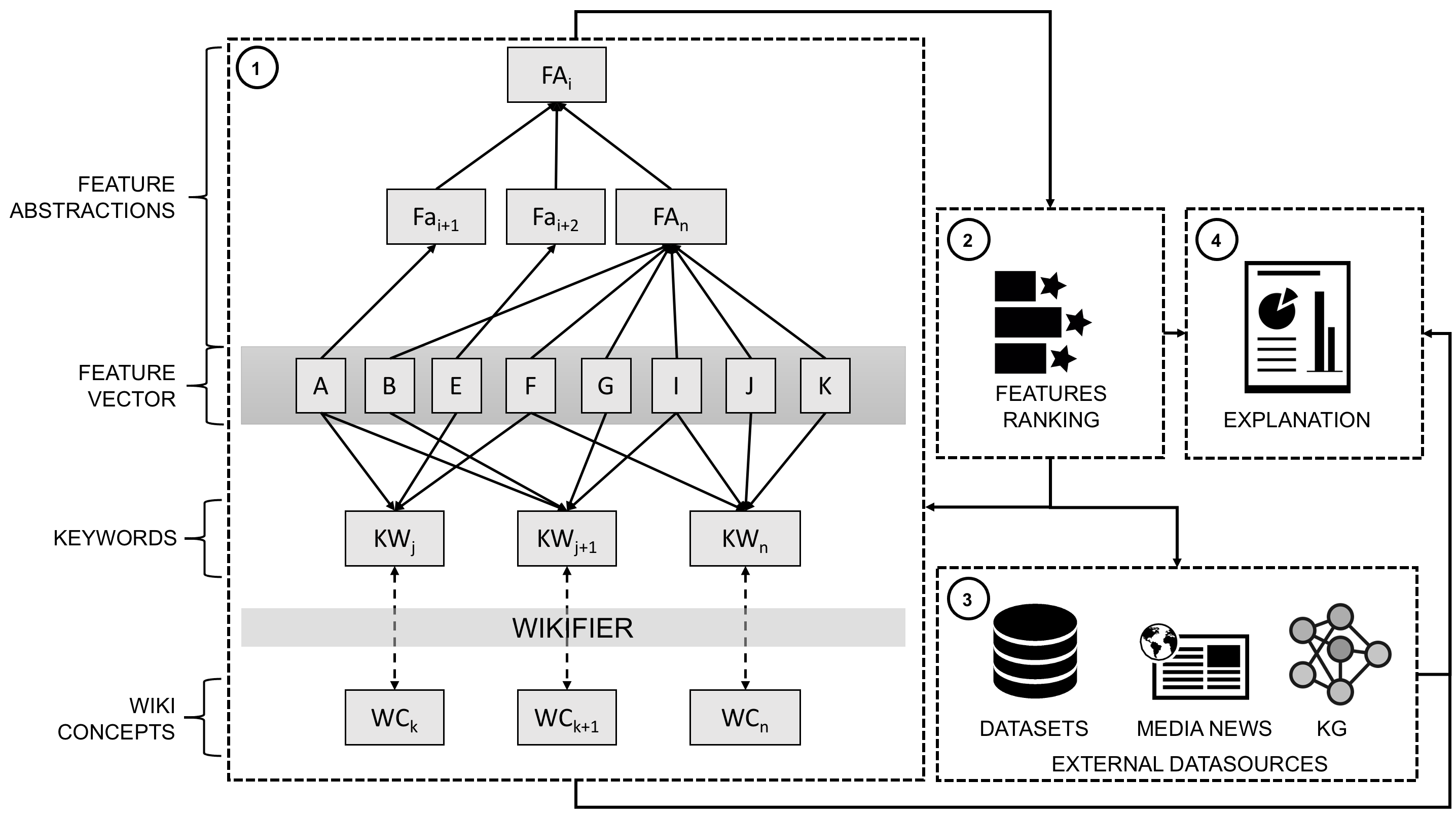}
\caption{High-level diagram of the components taken into account and the procedure followed to craft the explanations.}
\label{F:XAI-FLOW}
\end{figure*}

This research was developed for a demand forecasting use case. It extends work performed at \cite{rovzanec2022knowledge}, and \cite{rovzanec2021xai}, taking into account four data sources: (i) data provided by a European original equipment manufacturer targeting the global automotive industry market; (ii) news media entries provided by Event Registry \cite{leban2014event}, a well-established Media Events Retrieval System that provides real-time mainstream media monitoring; (iii) the EU Open Data Portal \cite{EUOpenDataPortal}; and (iv) the Google Knowledge Graph (Google KG) \cite{noy2019industry}. We detail the procedure in Fig. \ref{F:XAI-FLOW}.

\begin{equation}\label{E:JACCARD_INDEX}
    JaccardIndex(A,B) = \frac{|\textit{A}\cap\textit{B}|}{|\textit{A}\cup\textit{B}|}
\end{equation}

\begin{equation}\label{E:JACCARD_DISTANCE}
    JaccardDistance = 1-JaccardIndex
\end{equation}

In \cite{rovzanec2022knowledge} we studied how explanations given to the user could be enriched with external sources, searching for news media entries and metadata regarding datasets. Searches were performed based on keywords that described features' semantic abstractions and new keywords identified from retrieved news media entries. The entries were then ranked based on the word movers distance (\cite{kusner2015word}) between embeddings. In this research, we adopted a different strategy and (i) computed wiki concepts for the features' abstraction keywords (see Table \ref{T:WIKIFICATION-MAPPINGS}), (ii) ranked news media entries based on the Jaccard distance (see Eq. \ref{E:JACCARD_DISTANCE}) between relevant reference wiki concepts and the ones obtained by wikifying the news media events \cite{brank2017annotating}, (iii) queried and ranked external data sources based on reference wiki concepts and the most important concepts that emerged from news media events, and (iv) further enriched the explanations by adding most relevant entries from the Google KG queried with the most relevant wiki concepts obtained from news media events. However, we found that most frequent wiki concepts in media news referred to persons and places, which were not informative to the task at hand. Therefore, we decided to filter them out to enhance the quality of the outcomes.

\begin{table*}[ht!]
\centering
\resizebox{0.50\textwidth}{!}{
\begin{tabular}{|l|l|}
\hline
\textbf{Feature keywords} & \textbf{Wiki concepts} \\ \hline
\multirow{2}{*}{Car Sales Demand} & Car \\ \cline{2-2} 
 & Demand \\ \hline
\multirow{2}{*}{New Car Sales} & Car \\ \cline{2-2} 
 & Sales \\ \hline
Vehicle Sales & Vehicle \\ \hline
\multirow{2}{*}{Car Demand} & Car \\ \cline{2-2} 
 & Demand \\ \hline
Automotive Industry & Automotive Industry \\ \hline
\multirow{2}{*}{Global GDP Projection} & Gross Domestic Product \\ \cline{2-2} 
 & Gross World Product \\ \hline
\multirow{2}{*}{Global Economic Outlook} & Economy \\ \cline{2-2} 
 & World economy \\ \hline
\multirow{2}{*}{Economic Forecast} & Forecasting \\ \cline{2-2} 
 & Economy \\ \hline
Unemployment Rate & Unemployment \\ \hline
Unemployment Numbers & Unemployment \\ \hline
Unemployment Report & Unemployment \\ \hline
Employment Growth & Employment \\ \hline
Long-term Unemployment & Unemployment \\ \hline
Purchasing Managers' Index & Manager (Gaelic games) \\ \hline
\end{tabular}
\caption{Mapping between feature keywords and wiki concepts. Wikification was performed invoking the service available at \url{https://wikifier.org}. Note that while most concepts were accurate, the wikification of the \textit{Purchasing Managers' Index} produced a wrong concept.
\label{T:WIKIFICATION-MAPPINGS}}}
\end{table*}

\section{Evaluation and Results}\label{RESULTS-AND-EVALUATION}

\begin{table*}[ht!]
\resizebox{0.95\textwidth}{!}{
\begin{tabular}{|l|l|r|r|}
\hline
\textbf{} & \textbf{Metric} & \multicolumn{1}{l|}{\textbf{Embeddings-based approach}} & \multicolumn{1}{l|}{\textbf{Semantics-based approach}} \\ \hline
 & Average Precision@1 & \textbf{0,97} & 0,95 \\ \cline{2-4} 
 & Average Precision@3 & \textbf{0,97} & 0,95 \\ \cline{2-4} 
 & RDE@1 & 0,30 & \textbf{0,38} \\ \cline{2-4} 
\multirow{-4}{*}{\textbf{Media Events}} & RDE@3 & 0,11 & \textbf{0,14} \\ \hline
 & Average Precision@1 & \textbf{0,77} & 0,71 \\ \cline{2-4} 
 & Average Precision@3 & \textbf{0,78} & 0,72 \\ \cline{2-4} 
 & RDE@1 & \textbf{0,14} & 0,01 \\ \cline{2-4} 
\multirow{-4}{*}{\textbf{Media Events' K\&WC}} & {\color[HTML]{333333} RDE@3} & \textbf{0,09} & 0,01 \\ \hline
 & {\color[HTML]{333333} Average Precision@1} & 0,56 & \textbf{0,68} \\ \cline{2-4} 
\multirow{-2}{*}{\textbf{External Datasets}} & {\color[HTML]{333333} RDE@1} & 0,41 & \textbf{0,43} \\ \hline
 & {\color[HTML]{333333} Average Precision@1} & NA & \textbf{0,76} \\ \cline{2-4} 
 & Average Precision@3 & NA & \textbf{0,46} \\ \cline{2-4} 
 & RDE@1 & NA & \textbf{0,15} \\ \cline{2-4} 
\multirow{-4}{*}{\textbf{Google KG}} & RDE@3 & NA & \textbf{0,09} \\ \hline
\end{tabular}
\caption{Results we obtained by analyzing the forecast explanations created for 56 products over three months. The best results are bolded. \textit{Media Events}, \textit{Media Events' Keywords}, \textit{External Datasets} and \textit{Google KG} correspond to contextual information displayed for each forecast explanation. \textit{Media Events' K\&WC} stands for Media Events' Keywords and Wiki Concepts.\label{T:RESULTS}}}
\end{table*}

Our primary interest in this research was to use semantic technologies and external data sources to enrich the explanations with (a) contextual information (e.g., events informed in media news that could explain a given forecast), (b) datasets' metadata that could serve to future model enrichments, and (c) new relevant concepts obtained from integrations with semantic tools and knowledge graphs. We realized (c) for this research by retrieving wiki concepts associated with media news pieces of interest and the results obtained from the Google KG.

\begin{equation}\label{E:RDE}
    RDE = \frac{\textit{Unique Entries}}{\textit{Total Listed Entries}}
\end{equation}

To evaluate the outcomes of such enrichment, we used two metrics to assess whether the entries were precise and diverse: (a) Average Precision@K, and the Ratio of Diverse Entries (RDE@K) (see Eq.~\ref{E:RDE}). The first metric allowed us to measure how much of the information we displayed was related to the underlying model's features. Given that the users rely on trusted automation \cite{lee2004trust}, we consider that exact results are required to increase the users' trust in the underlying application. Furthermore, considering that curiosity is related to the workers' engagement, we consider a diverse set of entries is preferred to foster it. We quantify such diversity through the RDE@K metric. While the entries do not repeat themselves in a single forecast explanation, nothing prevents having the duplicate entries listed through different forecast explanations. We consider that the best case could avoid repeated entries to maximize users' learning. Nevertheless, different strategies could be adopted, framing the entries display as a recommender system problem and considering users' implicit and explicit feedback to rank and decide whether and where to display them \cite{kilani2018using,karimi2018news,sidana2021user}.

We present our results in Table \ref{T:RESULTS}. In the Table we compare results obtained from our previous work (\textit{Embeddings-based approach}) \cite{rovzanec2022knowledge} and the \textit{Semantics-based approach} described in Section \ref{USE-CASE}. We found that the former approach had a better performance when considering the media events' diversity, trading a slight decay in precision compared with the embeddings-based approach. Regarding the Media Events' K\&WC, the embeddings-based approach achieved better for both diversity and precision. While differences in precision were contested, they were pronounced regarding diversity. We consider it is natural to have a less numerous set of wiki concepts compared to keywords. Nevertheless, we consider that the metric values could be improved. On the other hand, the semantics-based approach was much more precise when recommending datasets and displayed a slightly higher diversity. Finally, when evaluating the entries related to the Google KG, we observed that the diversity was similar to the one obtained for media news with the embeddings-based approach. The first results were precise, but the precision dropped by 0,30 points when considering k=3. We found that those considered erroneous were mostly related to the economy or automotive industry when analyzing the entries. Nevertheless, they were not useful for the explanations at hand since they referenced banks (e.g., Bank of America or Deutsche Bank) or prominent figures (e.g., Edward Fulton Denison, who pioneered the measurement of the United States' Gross Domestic Product, or Kathleen Wilson-Thompson, independent director at Tesla Motors Inc. at the time of this writing). On the other hand, we considered the entries were accurate when they referenced companies from the automotive sector (e.g., Faurecia, Rivian Automotive Inc., Polestar or Vinfast) or related to it (e.g., \textit{Plug Power} which develops hydrogen fuel cell systems to replace conventional batteries or the \textit{Flinkster} carsharing company).

Finally, considering the results regarding Media Events' K\&WC and external datasets, we consider that the embeddings-based approach should be used to retrieve new keywords and concepts displayed to the users. In contrast, the semantics-based approach provides the best concepts that lead to better results when searching for external datasets' metadata.

\section{Conclusions and Future Work}\label{CONCLUSION}
Along with the increasing adoption of Artificial Intelligence in manufacturing, explainability techniques must be developed to ensure users' can learn the models' behaviors. Furthermore, explanations provided to the user can be enriched with additional insights that foster the users' curiosity, resulting in an exploratory dynamic towards the artificial intelligence application and domain-specific problems and enabling the development of trust towards such applications. One way to achieve this goal is to enrich the explanations with information obtained from external sources to augment users' knowledge and help them make responsible decisions. 

This research explored augmenting explanations by incorporating media news, datasets' metadata, and information queried from open knowledge graphs. Furthermore, we compared two approaches (based on embeddings and wiki concepts) to rank the data sources' entries and retrieve new concepts and keywords from them. We found that results were similar when considering media events but differed for new keywords and concepts (embeddings-based approach was best) and external datasets (semantic-based approach was best). Finally, while the integration with the Google KG proved informative, new strategies must be explored to increase the precision of results when showing multiple entries.

In future work, we would like to explore how critical components of an explanation encoded in a graph structure can be used to enhance the explanations and whether users' feedback can lead to better explanations if we frame them as a recommender system problem.

\vspace{6pt} 



\authorcontributions{Conceptualization, J.M.R.; methodology, J.M.R.; software, E.T. and J.M.R.; validation, J.M.R.; formal analysis, J.M.R.; investigation, J.M.R.; resources, J.M.R., P.Z. and E.T.; data curation, J.M.R.; writing---original draft preparation, J.M.R.; writing---review and editing, J.M.R. and K.K.; visualization, J.M.R.; supervision, D.M. and B.F.; project administration, K.K and I.N. ; funding acquisition, K.K, B.F. and D.M. All authors have read and agreed to the published version of the manuscript.}

\funding{This work was supported by the Slovenian Research Agency and the European Union’s Horizon 2020 program projects FACTLOG under grant agreement H2020-869951, and STAR under grant agreement number H2020-956573.}



\dataavailability{No data was released.} 

\acknowledgments{This document is the property of the STAR consortium and shall not be distributed or reproduced without the formal approval of the STAR Management Committee. The content of this report reflects only the authors' view. The European Commission is not responsible for any use that may be made of the information it contains.}




\abbreviations{Abbreviations}{
The following abbreviations are used in this manuscript:\\

\noindent 
\begin{tabular}{@{}ll}
Google KG & Google Knowledge Graph \\
LACE & Local Agnostic attribute Contribution Explanation \\
LIME & Local Interpretable Model-agnostic Explanations \\
LoRE & LOcal Rule-based Explanation \\
Media Events' K\&WC & Media Events' Keywords and Concepts \\
RDE & Ratio of Diverse Entries \\
SHAP & SHapley Additive exPlanations \\
XAI & Explainable Artificial Intelligence
\end{tabular}}


\begin{adjustwidth}{-\extralength}{0cm}

\reftitle{References}


\bibliography{main.bib}


%


\end{adjustwidth}
\end{document}